\pdfoutput=1
\documentclass[11pt]{article}

\usepackage{acl}

\usepackage{times}
\usepackage{latexsym}

\usepackage[T1]{fontenc}
\usepackage[utf8]{inputenc}

\usepackage{microtype}

\PassOptionsToPackage{table}{xcolor}
\usepackage{microtype}
\usepackage{graphicx}
\usepackage{listings}
\usepackage{xcolor}

\usepackage{comment}

\usepackage{graphicx}

\title{FurChat: An Embodied Conversational Agent using LLMs, Combining Open and Closed-Domain Dialogue with Facial Expressions}
\author{Neeraj Cherakara, {\bf Finny Varghese}, {\bf Sheena Shabana}, {\bf Nivan Nelson} \\ {\bf Abhiram Karukayil}, {\bf Rohith Kulothungan}, {\bf Mohammed Afil Farhan}, \\ {\bf Birthe Nesset}, {\bf Meriam Moujahid}, {\bf Tanvi Dinkar}, {\bf Verena Rieser}, {\bf Oliver Lemon}$\dagger$  \\
  Interaction Lab, Heriot-Watt University ; $\dagger${Alana AI}\\
  \texttt{\{nc2025, fv2002, ss2022, nn2023, ak2120, rk2065, mf2034, bn25} \\
  \texttt{mm470, t.dinkar, v.t.rieser, o.lemon\}@hw.ac.uk} \\
  }

\begin{document}
\maketitle
\begin{abstract}
We demonstrate an embodied conversational agent that can function as a receptionist and generate a mixture of  open and closed-domain dialogue along with facial expressions, by  using a large language model (LLM) to develop an engaging conversation. We deployed the system onto a Furhat robot, which is highly expressive and capable of using both verbal and nonverbal cues during interaction. The system was designed specifically for the National Robotarium to interact with visitors through natural conversations, providing them with information about the facilities, research, news, upcoming events, etc. The system utilises the state-of-the-art GPT-3.5 model to generate such information along with domain-general conversations and facial expressions based on prompt engineering.
%Two systems have been developed, one of which generates closed-domain dialogues, and the other generates open-domain dialogues. The closed-domain system employs a formal style of conversation with no facial expressions, whereas the open-domain system employs an informal style of conversation with appropriate facial expressions.

\end{abstract}

\section{Introduction}
%Robotics and artificial intelligence advancements over the past few decades have resulted in the development of robots for applications other than their traditional roles in industries.

The progress in robotics and artificial intelligence in recent decades has led to the emergence of robots being utilized beyond their conventional industrial applications. Robot receptionists are designed to interact with and assist visitors in various places like offices, hotels, etc. by providing information about the location, services, and facilities. The appropriate use of verbal and non-verbal cues is very important for the robot's interaction with humans \citep{MAVRIDIS201522}. Most research in the field has been mainly focused on developing domain-specific conversation systems, with little exploration into open-domain dialogue for social robots.

% \begin{comment}
\begin{figure}
\centering
\includegraphics[width=0.48\textwidth]{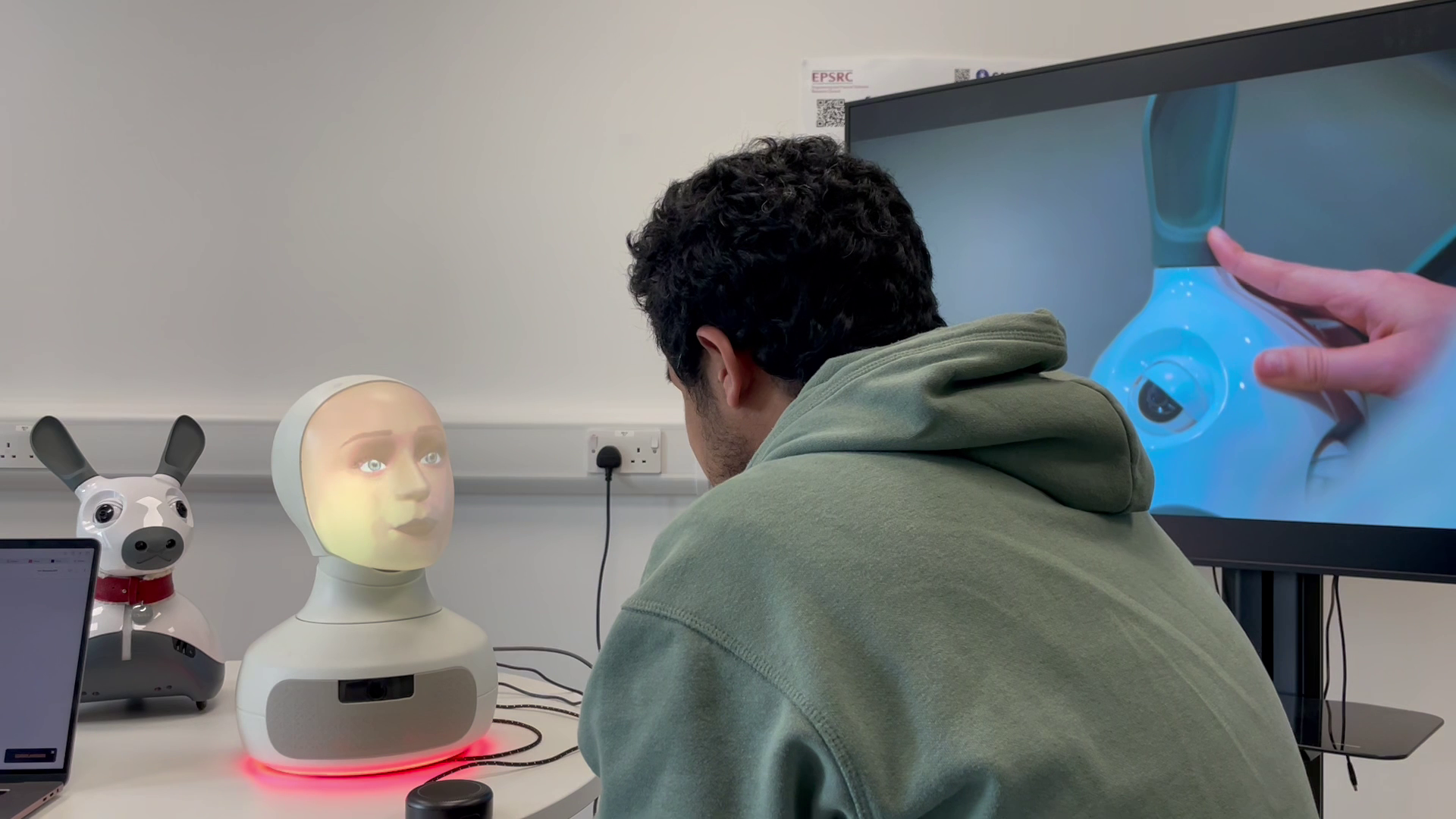}
\caption{A user interacting with the FurChat System.}
% To watch or see the video, click on the link: \url{https://youtu.be/fwtUl1kl22s} }
\label{fig:interface}
\end{figure}
% \end{comment}

\begin{figure*}[ht!]
  \centering
  \includegraphics[width=0.8\linewidth]{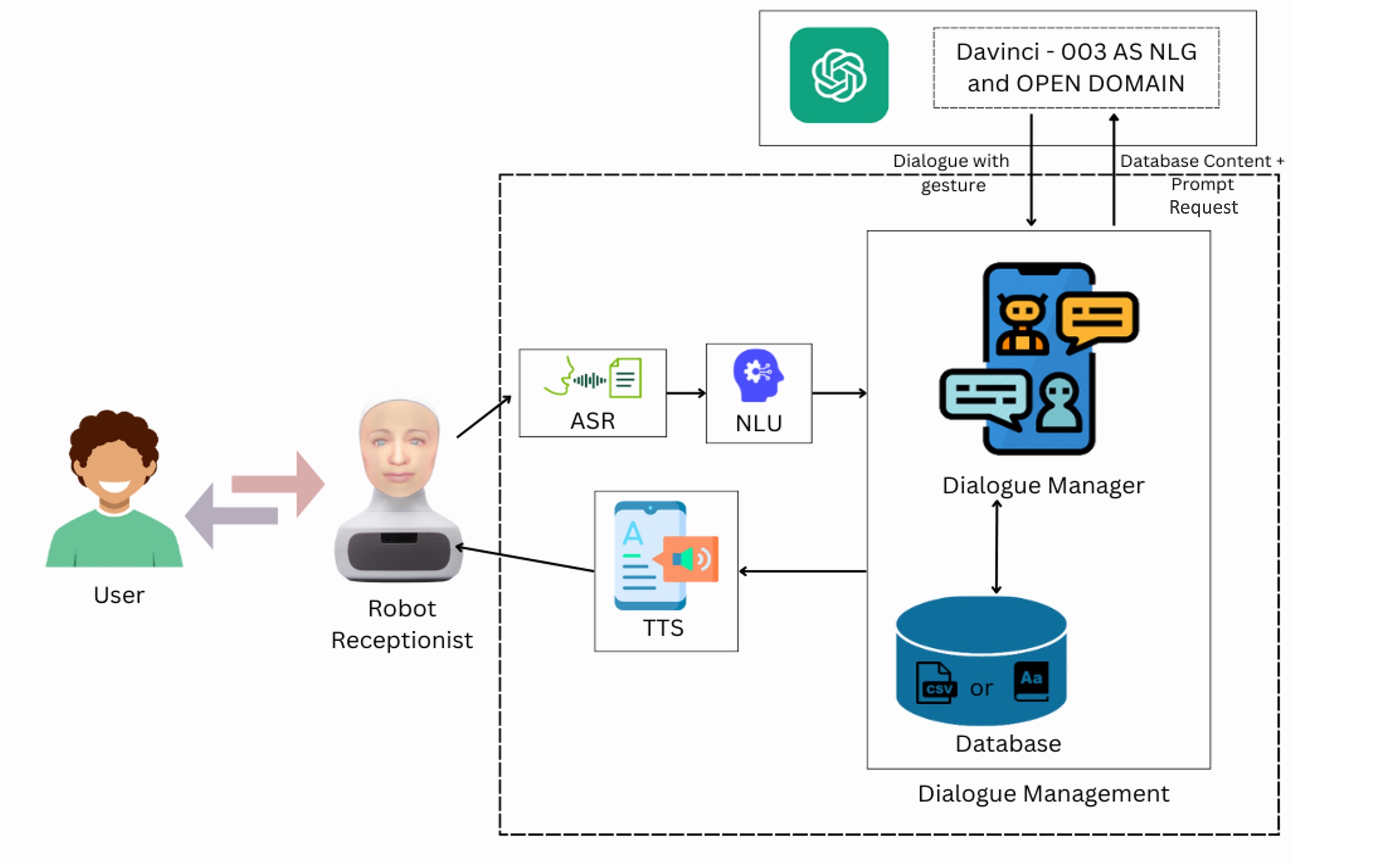}
\caption{System Architecture of the current FurChat system.}
\label{fig:system_arch}
\end{figure*}

 Conventional agents are often rule-based, which means they rely on pre-written commands and keywords that are pre-programmed. This limits the interaction with humans to little or no freedom of choice in answers \citep{info:doi/10.2196/17158}. The advancement of large language models (LLMs) in the past year has brought an exciting revolution in the field of natural language processing. With the development of  models like 
 % GPT-3\footnote{\url{https://platform.openai.com/docs/models/gpt-3}}, 
 GPT-3.5\footnote{\url{https://platform.openai.com/docs/models/gpt-3-5}}, 
 % which was developed by OpenAI\footnote{\url{https://openai.com/}}
 we have seen unprecedented progress in tasks such as question-answering and text summarization \citep{brown2020language}. However, a question remains about how to successfully leverage the capabilities of LLMs to create systems that can go from closed domain to open, while also considering the embodiment of the system.

In this work, we present FurChat\footnote{A demonstration video of the system is available \href{https://youtu.be/fwtUl1kl22s}{here}.}, an embodied conversational agent that utilises the latest advances in LLMs to create a more natural conversational experience. The system seamlessly combines open and closed-domain dialogues with emotive facial expressions, resulting in an engaging and personalised interaction for users. The system was initially designed and developed to serve as a receptionist for the National Robotarium, in continuation of the multi-party interactive model developed by \citet{moujahid2022demonstration}, and its deployment shows promise in other areas due to the LLMs versatile capabilities. As a result, the system is not limited to the designated receptionist role, but can also engage in open-domain conversations, thereby enhancing its potential as a multifunctional conversational agent. We demonstrate the proposed conversational system on a Furhat robot \citep{al2013furhat} which is developed by the Swedish firm Furhat Robotics\footnote{\url{https://furhatrobotics.com/}}. With FurChat, we demonstrate the possibility of LLMs for creating a more natural and intuitive conversation with robots.

\section{Furhat Robot}
%Furhat Robotics has designed a social robot, Furhat, which uses advanced conversational AI and expressive facial expressions to communicate naturally and intuitively with humans. The robot features a three-dimensional mask that projects an animated face\cite{al2013furhat}, mimicking a human's facial expressions, and is supported by a motorized platform that allows its head to spin and nod. Additionally, Furhat is equipped with a microphone array and speakers to recognize and respond to human speech.
Furhat is a social robot created by Furhat Robotics. To interact with humans naturally and intuitively, the robot employs advanced conversational AI and expressive facial expressions. A three-dimensional mask that mimics a human face is projected with an animated face using a microprojector \cite{al2013furhat}. A motorised platform supports the robot's neck and head, allowing the platform's head to spin and node. To identify and react to human speech, it has a microphone array and speakers. 
Due to the human-like appearance of Furhat, it is prone to the uncanny valley effect \cite{aagren2022exploring}.

\section{System Architecture}
\label{sys}
As shown in Figure \ref{fig:system_arch}, the system architecture represents a conversational system that enables users to interact with a robot through spoken language. The system involves multiple components, including automatic speech recognition (ASR) for converting user speech to text, natural language understanding (NLU) for processing and interpreting the text, a dialogue manager (DM) for managing the interaction flow, and natural language generation (NLG) powered by GPT-3.5 for generating natural sounding responses\citep{10.1145/3581641.3584037}. 
% with facial expressions. 
The generated text is then converted back to speech using text-to-speech (TTS) technology and played through the robot's speaker to complete the interaction loop. The system relies on a database to retrieve relevant data based on the user's intent.

%The proposed system architecture is shown in the Figure \ref{fig:system_arch}. 
%When the user is engaged in a conversation with the robot, the inbuild Automatic Speech Recognition (ASR) module of Furhat will be activated. The data received from the ASR will be integrated with the in-build Natural Language Understanding (NLU) module of the robot, which will identify the content of the speech and will aid the Dialogue Manager (DM) to identify the indents to be triggered. When an indent in the NLU is triggered, it sends a key value to the local Natural Language Generation (NLG) Database prepared with the custom made contents as per the application. The DM module then requests the web based Large Language Model (LLM), i.e GPT to create dialogues based on the user's conversation and the obtained description from the database. The LLM reply back the DM with the generated dialogue, which will be converted to speech by the furhat Text to Speech (TTS) module. During the communication with DM and LLM, the DM will prompt the model to generate emoticons based on the ongoing conversation, which will be used by the DM to generate facial expressions along with the ongoing dialogues.

\subsection{Speech Recognition}
The current system uses the Google Cloud Speech-to-Text\footnote{\url{https://cloud.google.com/speech-to-text}} module for ASR. This module, which transcribes spoken words into text using machine learning algorithms, is integrated into the system by default through the Furhat SDK.
% \footnote{\url{https://docs.furhat.io/getting_started}}.

\subsection{Dialogue Management}
%NLU
%Database
%Flows and states
Dialogue Management consists of three sub-modules: NLU, DM and a database storage. The NLU component analyses the incoming text from the ASR module and, through machine learning techniques, breaks it down into a structured set of definitions \cite{9075398}. The FurhatOS provides an NLU model to classify the text into intents based on a confidence score. We provide multiple custom intents
% with example human responses 
for identifying closed-domain intents using Furhat's NLU capabilites.

The in-built dialogue manager in the Furhat SDK is responsible for maintaining the flow of conversation and managing the dialogue state based on the intents identified by the NLU component. This module is responsible for sending the appropriate prompt to the LLM, receiving a candidate response from the model,
% the appropriate response from the model, and processing the output response to generate speech.
and subsequent processing of the response to add in desired facial gestures (see \S\ref{gesture}).

%There is a custom database, which is created using web-scrapping to generate the content for conversation on the specific application. This is a pre-written data base dictionary, which has the intends as the keys and the respective scrapped data as the value, which act as the reference descriptor for the dialogue generation. The database will be accessed by the main state, which will then trigger the GPT3DialogueGenerator for dialogue generation. If the user asks anything outside the context of the application, the database will not accessed by the DM, and the conversation will be transfered to the LLM for dialogue generation without any specific reference content.

An open challenge faced by present-day LLMs is the \emph{hallucination of nonfactual content}, which potentially undermines user trust and raises concerns of safety.
% which happens mainly because of the unavailability of data during the training of the model. 
While we cannot fully mitigate hallucinated content in the generated responses, in order to tone-down this effect, we create a custom database following suggestions from \citet{kumar2023geotechnical}. We do so by manually web-scraping the website of the National Robotarium\footnote{\url{https://thenationalrobotarium.com/}}. The database consists of a dictionary of items with the intents as keys and scraped data as values. When an appropriate intent is triggered, the dialogue manager accesses the database to retrieve the scraped data, which is then sent with the prompt (further details in \S\ref{prompt}))
to elicit a 
% valid and accurate 
response from the LLM.

\subsection{Prompt engineering for NLG}
\label{prompt}
%Prompt engineering
%Davinci Integration
%Emoticons generation
The NLG module is responsible for generating a response based on the request from the dialogue manager. Prompt engineering is done to elicit an appropriate sounding response from the LLM, which generates natural dialogue that results in engaging conversations with humans. The current system uses {\tt text-davinci-003}, which is one of the most powerful models in the GPT-3.5 series and it is priced at \$0.0200 per 1000 tokens.

% This model can take an input of about 4097 tokens as it is updated until June 2021.
Producing relevant responses was achieved using the combined technique of few-shot learning and prompt engineering, which enabled us to try different variations in techniques and produce a variety of output by the LLM.
\begin{comment}    
NLG module handles the dialogue generation based on the request raised by the DM. In the NLG module prompt engineering is used to ask the right question of the LLM, which will create a natural dialogue,  resulting in an engaging conversation. NLG module store the dialogue history of past conversation, to generate responses in the context of ongoing conversation. In the proposed system Davinci-003 version is used to integrate GPT-3 with that of this model.But this architecture can be updated with any advanced LLMs for better performance. 

[\emph{This is a conversation with a robot receptionist}, <Robot personality>, <Information>, <Dialogue history>, <Response Format along with sample emoticons>],
\end{comment}

%During prompt engineering, the personality of the robot and the context of the application are described, along with the past few dialogue histories and a response format. 
During prompt engineering, the personality of the robot and the context of the application are described, along with the past few dialogue histories and scraped data from the database in a particular response format.
% This process ensures that the conversation has the required nature and duration.
Moreover, the prompt engineering methodology involves using the LLM to generate an appropriate emoticon based on the conversation. In the context of emotional expression during an interaction, selecting an appropriate emoticon depends on understanding the underlying emotions being conveyed by the visitors and adhering to the display rules of the specific social situation. If the dialogue reflects joy or humor, a happy facial gesture might be fitting. On the other hand, if the conversation conveys empathy or sadness, a sad face could be more suitable. These emoticons are then integrated with the robot's facial gestures to generate facial expressions (see \S\ref{gesture}), thereby 
% making the conversation more engaging and expressive. 
enabling a text-based LLM to integrate in the embodied Furhat robot.
% By explicitly specifying the personality and context in the prompt will ensure that the conversation between the robot and the human is coherent and relevant to the topic.
The explicit specification of the personality and context in the prompt aids in creating a natural conversation between the robot and the human that is coherent and relevant to the topic. The sample format of the prompt used is as follows:

\emph{This is a conversation with a robot receptionist, <Robot Personality>, <Data from the Database>, <Dialogue history>, <Response Format along with sample emoticons>}.

\subsection{Gesture Parsing}
\label{gesture}
%Gesture identification
%Gesture Placement
The Furhat SDK offers a range of built-in facial gestures that can be enhanced by custom facial gestures that meet specific needs. The latest GPT models have the ability to recognise emotions and sentiments from text, which is used in the  system \cite{leung2023application}. Rather than simply recognising sentiments in the text, the model is tasked with generating appropriate emotions for the conversation from the text
% , resulting in a more emotionally expressive and engaging dialogue. 
After receiving the response from the model, the matched conditional clause in the dialogue manager will trigger an expression from the pre-developed set of gestures, which will be triggered along with the generated speech.

%Gestures are generated based on the ongoing dialogues by converting emoticons received from the LLM into in-built and custom made furhat gestures. Gestures are placed in the context through prompt engineering in NLG, which will raise a request for gesture based in the emoticons received in the DM. The matched conditional clause in the DM will trigger an expression from the pre-developed set of gestures, which will be triggered along with the generated speech. 

\subsection{Text-To-Speech Generation}
For converting the text to speech, the Amazon Polly\footnote{\url{https://docs.aws.amazon.com/polly/}} service is used. This service is available within FurhatOS by default. 
% The speech is then sent to the installed speaker on the Furhat robot.

\section{Example Dialogue}

%A sample conversation between the robot and human are shown in the Table \ref{table:ex_d1}:
An example dialogue between a human and the robot is illustrated in \autoref{table:ex_d1}.

\begin{table}[ht]
\renewcommand{\arraystretch}{1.2}
\centering
\small
\begin{tabular}{p{7cm}}
\textit{At the outset, the robot remains in an idle state. Once the user enters the vicinity of the robot, the conversation begins.} \\
%R: \textcolor{blue}{[reception\_bot]} Hi, I'm Ari, how can I help you?\\
R: \textcolor{blue}{[robot]} Hello, I am the Receptionist here at the National Robotarium. Would you like to know about this facility? \textcolor{purple}{<Robot smiles>} \\
%\textit{Robot Smiles} \\
U: \textcolor{gray}{[user]} Yes, tell me about this facility.\\
%\textit{The intent \emph{facility} is then searched in the database to fetch the details of the facilities available. This information is sent along with the prompt to the text-davinci-003 model to generate dialogue with emoticons. Then the response is processed and converted to speech.}\\
R: \textcolor{blue}{[robot]} Sure. \textcolor{purple}{<Robot smiles>}\\
%\textit{Robot Smiles}\\
R: \textcolor{blue}{[robot]} The National Robotarium is a state-of-the art research facility located on the Heriot-Watt campus in Edinburgh. \textit{\dots} \\
%It was inaugurated in September 2022 and is collaboration between the University of Edinburgh and Heriot-Watt. It mainly focuses on robotics and AI related projects and researches. \\
U: \textcolor{gray}{[user]} That's great. Is there any ongoing healthcare researches here \\
R: \textcolor{blue}{[robot]} Yes. \textcolor{purple}{<Robot smiles>} The National Robotarium is part of a multimillion-pound collaborative project called SPRING and is focused on developing socially assistive robots for healthcare settings. SPRING will focus on supporting elderly patients and improving psychological well-being and the relationship between patients and hospital professionals. \textcolor{purple}{<Robot nod>}\\
% \textit{Robot gives a small head nod} \\
%\textit{\dots}\\
U: \textcolor{gray}{[user]} That's nice. Can you tell a joke?\\
%\textit{Robot smiles}\\
R: \textcolor{blue}{[robot]} \textcolor{purple}{<Robot smiles>} Sure, Why did the robot go to the bar? \textcolor{purple}{<Robot winks>} To get a screwdriver!\\
U: \textcolor{gray}{[user]} Can you suggest me a movie about robots?\\
R: \textcolor{blue}{[robot]} Sure thing! \textcolor{purple}{<Robot smiles>} How about "I, Robot"? It's a classic that deals with the exploration of robotics and AI. \textcolor{purple}{<Robot nod>}
%\textit{Robot winks}\\
%R: \textcolor{blue}{[robot]} To get a screwdriver!\\
% \textit{Robot gives a long smile}\\
% U: \textcolor{gray}{[user]} That is a good joke. Thank you. I need to go. Goodbye.\\
% R: \textcolor{blue}{[robot]} No problem! It was nice talking to you!\\
% \textit{Robot smiles}\\
% R: \textcolor{blue}{[robot]} Have a great day!\\

\end{tabular}
\caption{Sample Conversation between the user and the robot. For a full system description, please refer to \S\ref{sys}.}
\label{table:ex_d1}
\end{table}

\begin{comment}
    
\paragraph{Initial State - } Robot is in idle state until the user is detected, and once user enter the conversation WhatCanIDo state and main state will be activated and robot will greet the user followed by listening to user's reply.

\noindent
\emph{\textbf{Robot response:}}\emph{"Hi, Would you like to know about this facility?"}

\noindent
\emph{\textbf{User response:}} \emph{"Tell me about this facility"}

\paragraph{Dialogue Management - } Dialogue is identified using ASR and ALU, the key \emph{"facility"} is then searched in the database to fetch details about the facility. This information returned from database will be used in prompt engineering to perform the dialogue generation. The response received from GPT3 is as follows \emph{"Sure, :) This facility is located in Edinburgh and was found in 2022"} and robot will generate dialogue and expressions based on the pre-built conditional check in DM.

\paragraph{Robot response: } 

\noindent
\emph{"Sure"} <express happiness gesture>
\emph{"This facility is located in Edinburgh and was found in 2022"}

Robot resets to main state and continue to listen.
\end{comment}

\section{Conclusions and Future Work}
We demonstrate FurChat, an embodied conversational agent with open and closed domain dialogue generation and facial expressions generated through LLMs, on a social robot in a receptionist environment. The system is developed by integrating the state-of-the-art GPT-3.5 model on top of the Furhat SDK. The proposed system uses a one-to-one interaction method of communication with the visitors. We plan on extending the system to handle multi-party interaction \cite{10.5555/3523760.3523907, addlesee2023data, lemon2022conversational, gunson2022developing}, which is an active research topic in developing receptionist robots. It is also crucial to address the issue of hallucination from the large language model and this problem can be mitigated by fine-tuning the language model and directly generating conversations from it without relying on any NLU components which we plan to implement in the future.

We plan to showcase the system on the Furhat robot during the SIGDIAL conference to all the attendees and show them the capabilities of using LLMs for dialogue and facial expression generation as described in this paper.

% %Review versions must not include acknowledgements.
\section*{Acknowledgements}
% \textit{To be added: SPRING project and Gender Bias}
This research has been funded by the EU H2020 program under grant agreement no.\ 871245 (\url{http://spring-h2020.eu/}) and the EPSRC project `Gender Bias in Conversational AI' (EP/T023767/1). 
% \begin{quote}
% \begin{verbatim}
% \bibliographystyle{references}
% \bibliography{custom}
% \end{verbatim}
% \end{quote}

\bibliography{references}
\bibliographystyle{acl_natbib}

% \appendix

% \section{Example Appendix}
% \label{sec:appendix}

% This is an appendix.

\end{document}